\documentclass[preprints,article,accept,moreauthors,pdftex]{Definitions/mdpi} 

\pdfoutput=1

\usepackage{float}
\usepackage{gensymb}
\usepackage{algorithm}
\usepackage{algorithmic}
\usepackage{footnote}
\usepackage{comment}
\usepackage{arydshln}
\history{}
\usepackage{amsmath,amssymb}

\usepackage[labelformat=simple]{subcaption} 

\DeclareCaptionLabelFormat{subcaptionlabel}{\normalfont(\textbf{#2}\normalfont)}
\usepackage{enumitem}

\def\bfR{{\bf R}}
\def\bfS{{\bf S}}
\def\bfW{{\bf W}}
\def\bfP{{\bf P}}
\def\bfA{{\bf A}}
\def\bfB{{\bf B}}
\def\bfM{{\bf M}}
\def\bfI{{\bf I}}
\def\bfC{{\bf C}}

\def\bfL{{\bf L}}

\def\bfE{{\bf E}}

\def\bfJ{{\bf J}}
\def\bfQ{{\bf Q}}

\def\bfF{{\bf F}}
\def\bfY{{\bf Y}}
\def\bfG{{\bf G}}

\newcommand{\norm}[1]{\left\lVert#1\right\rVert}
\def\bfR{{\bf R}}
\def\bfS{{\bf S}}
\def\bfW{{\bf W}}
\def\bfP{{\bf P}}
\def\bfA{{\bf A}}
\def\bfB{{\bf B}}
\def\bfM{{\bf M}}

\def\bfI{{\bf I}}
\def\bfC{{\bf C}}

\def\bfL{{\bf L}}

\def\bfE{{\bf E}}

\def\bfJ{{\bf J}}
\def\bfQ{{\bf Q}}

\def\bfF{{\bf F}}
\def\bfY{{\bf Y}}
\def\bfG{{\bf G}}

\Title{Structure from Articulated Motion: Accurate and~Stable Monocular 3D Reconstruction without Training Data}

\Author{{Onorina Kovalenko} $^{1,}$*, Vladislav Golyanik $^{2}$, Jameel Malik $^{1,3,4}$, Ahmed Elhayek $^{1,5}$ and~Didier~Stricker $^{1,3}$}

\AuthorNames{Onorina Kovalenko, Vladislav Golyanik, Jameel Malik, Ahmed Elhayek and Didier Stricker}

\address{%
$^{1}$ \quad Department Augmented Vision, German Research Center for Artificial Intelligence (DFKI), 67663~Kaiserslautern, Germany; jameel.malik@dfki.de (J.M.); ahmed.elhayek@dfki.de (A.E.); didier.stricker@dfki.de (D.S.) \\
$^{2}$ \quad Department of Computer Graphics, Max Planck Institute for Informatics, 66123~Saarbrücken, Germany; golyanik@mpi-inf.mpg.de \\
$^{3}$ \quad Department of Computer Science, University of Kaiserslautern, 67663~Kaiserslautern, Germany \\
$^{4}$ \quad School of Electrical Engineering and Computer Science (SEECS),
National University of Sciences and~Technology (NUST), 44000~Islamabad, Pakistan \\
$^{5}$ \quad Department of Computer Science, University of Prince Mugrin (UPM), 20012~Madinah, Saudi Arabia;} 

\corres{Correspondence: onorina.kovalenko@dfki.de}

\abstract{Recovery of articulated 3D structure from 2D observations is a challenging computer vision problem with many applications.
 Current learning-based approaches achieve state-of-the-art accuracy on public benchmarks but are restricted to specific types of objects and motions covered 
 by the training datasets. 
 Model-based approaches do not rely on training data but show lower accuracy on these datasets. 
 In this paper, we introduce a model-based method called \textit{{Structure from Articulated Motion}} (SfAM), which can recover multiple object and motion types without training on extensive data collections. 
 At the same time, it performs on par with learning-based state-of-the-art approaches 
 on public benchmarks and outperforms previous non-rigid structure from motion (NRSfM) methods. SfAM is built upon a general-purpose NRSfM technique while integrating a soft spatio-temporal constraint on the bone lengths.
We use alternating optimization strategy to recover optimal geometry ({i.e.,} bone proportions) together with 3D joint positions by enforcing the bone lengths consistency over a series of frames. 
 SfAM is highly robust to noisy 2D annotations, generalizes to arbitrary objects and does not rely on training data, which is shown in extensive experiments on public benchmarks and real video sequences. 
 We believe that it brings a new perspective on the domain of monocular 3D recovery of articulated structures, including human motion capture. }

\begin{document}

\section{Introduction}
3D structure recovery of articulated objects ({i.e.,} comprising multiple connected rigid parts) from a set of 2D point tracks through multiple monocular images is a challenging computer vision problem~\cite{Ramakrishna2012, Wandt2016,zhou2016sparseness,DBLP:conf/icra/LeonardosZD16}. Articulated structure recovery is ill-posed due to missing information about the~third dimension~\cite{Lee1985DeterminationO3}. 
Its applications include gesture and activity recognition, character animation in~movies and games, and motion analysis in sport and robotics. 

Recently, multiple learning-based approaches that recover 3D structures from 2D landmarks have been introduced~\cite{Hossain18,DBLP:conf/iccv/ZhouH0XW17,DBLP:conf/3dim/MehtaRCFSXT17,Martinez2017}. 
These methods show state-of-the-art accuracy across public benchmarks. However, they are restricted to a specific kind of structure ({e.g.,} human skeleton) and require extensive datasets for training. Moreover, they often fail to recover poses that are different from the training examples (see Section~\ref{sec:evalhumanyoutube}). 
When a scene includes different types of articulated objects, different methods have to be applied to reconstruct the whole scene.

\begin{figure}[H] 
\begin{center} 
     \includegraphics[width=13.6cm]{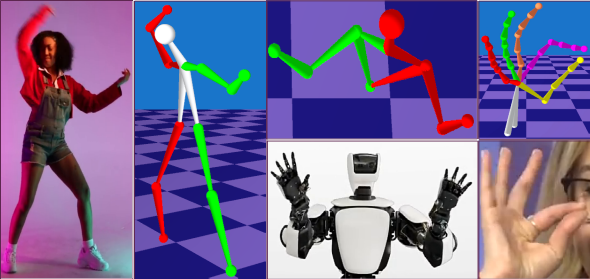}\\ 
   \caption{ We recover different articulated structures from real-world videos with high accuracy and~no~need for training data. 
   Our \textit{ Structure from Articulated Motion} ({SfAM}) 
approach is not restricted to a single object class and only requires a rough articulated structure prior. 
   The reconstructions are provided under different view~angles. 
   } 
\label{fig:examples} 
\end{center} 
\end{figure}

\begin{figure}[H]
 \captionsetup[subfigure]{labelformat=empty}
 \begin{center}
    \includegraphics[width=13.6cm]{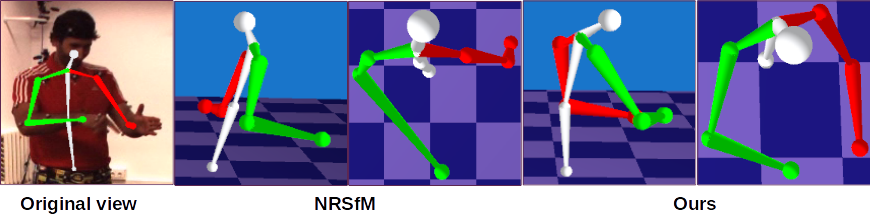}\\
    \caption{Side-by-side comparison of the non-rigid structure from motion ({NRSfM}) method~\cite{Ansari2017} and our SfAM. Reconstruction results of~\cite{Ansari2017} violate anthropometric properties of the human skeleton due to changing bone lengths from frame to~frame.}
    \label{fig:nrsfm}
    \end{center}
 \end{figure}

In this paper, we introduce a general approach for accurate recovery of 3D poses of any articulated structure from 2D observations that does not rely on training data (see Figure~\ref{fig:examples}).  
We build upon the~recent progress in non-rigid structure from motion (NRSfM), which is a general technique for non-rigid 3D reconstruction from 2D point tracks.  
However, when considering an articulated object as a general non-rigid one, reconstructions can evince significant variations in the distances between the connected joints  (see Section~\ref{sec:robustnessbonelengths}). These distances have to remain nearly constant across all articulated poses. Our method relies on this assumption and imposes a spatio-temporal constraint on~the~bone lengths. 
\vspace{-5mm}
We call our approach \textit{{Structure from Articulated Motion}} (SfAM). We apply an articulated structure term as a soft constraint on top of the classic optimization problem of NRSfM~\cite{Dai2014}. This term enforces the bone lengths---though not known in advance---to remain constant across all frames. Our~optimization strategy alternates between the classic NRSfM problem and our articulated structure term until they both converge. This allows for recovering the geometry together with the~3D joint positions and the method does not rely on known bone lengths.
Starting from a rough initialization of the articulated structure ({e.g.,} a human arm is longer than a leg), SfAM still converges to the correct structure proportions (see Section~\ref{sec:robustnessbonelengths}). 
Figure~\ref{fig:nrsfm} illustrates the significant difference between results produced by a general-purpose NRSfM technique~\cite{Ansari2017} and our SfAM. 

To summarise, our \textbf{{contributions}} are: 
\begin{itemize}[leftmargin=*,labelsep=5.8mm]
    \item \textbf{{A generic framework}} for articulated structure recovery which achieves state-of-the-art accuracy among not learning-based methods across public datasets. Moreover, it shows performance close to state-of-the-art learning-based methods but at the same time is not restricted to specific objects (see Section~\ref{sec:experiment}) and does not require training data.
    \item \textbf{{SfAM recovers sequence-specific bone proportions}} together with 3D joints (see Section~\ref{sec:method}). Thus,~it does need known bone lengths. 
    \item The \textbf{{articulated prior energy term}} makes our approach robust to noisy 2D observations (see~Section~\ref{sec:robustness2D}) by imposing additional constraints on the 3D structure.
\end{itemize}

In this paper, 
we show that a not learning-based approach can perform on par  with state-of-the-art learning-based methods and even outperform some of them in real-world scenes (see Section~\ref{sec:evalhumanyoutube}).  
We demonstrate the effectiveness of SfAM for the recovery of different articulated structures through extensive quantitative and qualitative evaluation on different datasets~\cite{Ionescu2014,Akhter2011,tompson14tog} and real-world scenes (see Section~\ref{sec:experiment}). To the best of our knowledge, our SfAM is the first NRSfM approach evaluated on~such comprehensive datasets as Human 3.6m~\cite{Ionescu2014} and NYU hand pose~\cite{tompson14tog}. As a side effect of our method, it can be used for precise articulated model estimation (generate personalized human skeleton rigs (see Section~\ref{sec:robustnessbonelengths})).
This contrasts a lot with most recent supervised learning approaches which require extensive labeled databases for training, and still, often fail when unfamiliar poses are observed (see~Section~\ref{sec:evalhumanyoutube}). 
Moreover, minor changes in the inputs lead to significant variations in the poses, which makes the results of learning-based methods very difficult or impossible to reproduce.

\section{Related Work}\label{sec:related_work}
\noindent
\textbf{{Rigid and Non-Rigid Structure from Motion}.} %
Factorization-based Structure from Motion (SfM) is a general technique for 3D structure recovery from 2D point tracks. 
An SfM problem is well-posed for rigid objects due to the rigidity constraint~\cite{TomasiKanade92}. 
Early extensions of Tomasi and Kanade's method~\cite{TomasiKanade92} for the non-rigid case rely on rank and orthonormality constraints~\cite{Bregler2000, Brand2005}. 
Subsequent methods investigated shape basis priors~\cite{Xiao2004}, temporal smoothness priors~\cite{Bartoli2008}, trajectory space constraints~\cite{Akhter2008} as well as such fundamental questions as shape basis uniqueness~\cite{HartleyVidal2008, Akhter2009}. 
More recent methods combine priors in the metric and trajectory spaces ~\cite{GotardoM2011}. To improve the reconstruction of stronger nonlinear deformations, Zhu {et al.}~\cite{Zhu2014} introduce unions of linear subspaces. Dai {et al.}~\cite{Dai2014} propose an NRSfM method with as few additional constraints as possible. 
Lately, the focus of NRSfM research is drawn to the problem of scalability~\cite{Ansari2017, Kumar2018}, {i.e.,} the consistent performance across different  scenarios and linear computational complexity in the number of points. 
Our SfAM is a scalable approach which builds upon the work of Ansari {et al.}~\cite{Ansari2017}. 
In contrast to~\cite{Ansari2017}, we recover articulated structures with higher~accuracy. 

\noindent
\textbf{{Articulated and Multibody Structure from Motion}.} 
Over the last few years, several SfM approaches for articulated motion recovery were proposed. 
Some of them relax the global rigidity constraint for multiple parts~\cite{Paladini2012, Costeira1998} so that each of the parts is constrained to be rigid. 
They can handle relatively simple articulated motions, as the segmentation and the structure composition are assumed to be unknown~\cite{Paladini2012}. 
As a result, these methods are hardly applicable to such complicated scenarios as human and hand pose recovery.  
Tresadern and Reid~\cite{Tresadern_Reid_2005}, Yan and Pollefeys~\cite{Yan_Pollefeys_2008} and Palladini~{et~al.}~\cite{Paladini2012} address the articulated case with two rigid body parts and detect a hinge joint. 
Later, an~approach with spatial smoothness and segmentation dealing with an arbitrary number of rigid parts was proposed by Fayad {et al.}~\cite{Fayad_2011}. 
Park and Sheikh~\cite{ParkS11} reconstruct trajectories given parent trajectories and known bone length, known camera, and root motion for each frame. Their objective is highly nonlinear and requires good initialization of trajectory parameters. In contrast, our method recovers sequence-specific bone proportions and does not rely on given bone lengths.
Next, Valmadre {et al.}~\cite{Valmadre2012} propose a dynamic-programming approach for the reconstruction of articulated 3D trees from input 2D joint positions operating in linear time. 
Multibody SfM methods reconstruct multiple independent rigid body transformations and non-rigid deformations in the same scene~\cite{Costeira1998, Kumar2017}. 
In contrast, our approach is more general as it imposes a soft constraint of articulated motion on top of classic NRSfM. 

\noindent
\textbf{{Piecewise and Locally Rigid Structure from Motion}.} 
Piecewise rigid approaches interpret the~structure as locally rigid in the spatial domain~\cite{pub10310,Taylor2010}. 
Several methods divide the structure into patches, each of which can deform non-rigidly~\cite{Fayad2010, Lee2016}. 
High granularity level of operation allows these methods to reconstruct large deformations as opposed to methods relying on linear low-rank subspace models~\cite{Fayad2010}. 
Rehan {et al.}~\cite{Rehan2014} penalize deviations between the bone lengths from the average distances between the joints over the whole sequence. 
This form of constraint does not guarantee a realistic reconstruction though, as it struggles to compensate for inaccurate 2D estimations or 3D inaccuracies in short time intervals. 

\noindent
\textbf{{Monocular 3D Human Body and Hand Pose Estimation}.} Bone length constraints are widely used in the single-view regression of 3D human poses. 
One of the early works in this domain operates on single uncalibrated images and imposes constraints on the relative bone lengths~\cite{Taylor2000}. 
It is capable of reconstructing a human pose up to scale. 
Later, an enhancement for multiple frames with bone symmetry and rigidity constraints (joints representing the same bone move rigidly relative to each other) was introduced by Wei and Chai~\cite{Wei2009}. Akhter and Black~\cite{Akhter2015} use a pose prior that captures pose-dependent joint angle limits. 
Ramakrishna {et al.}~\cite{Ramakrishna2012} use a sum of squared bone lengths term that can still lead to unrealistic poses. 
Wandt {et al.}~\cite{Wandt2016} constrain the bone lengths to be invariant. 
Their trilinear factorization approach relies on pre-trained body poses serving as a shape prior and transcendental functions modeling periodic motion peculiar to the human gait. 
An adaptation of this approach to hand gestures would require the  acquisition of a new shape prior. Wandt~{et~al.}~\cite{DBLP:conf/eccv/WandtAR18} constrain the sum of squared bone lengths of the articulated structure to be invariant throughout image sequence. However, the length of each bone can still vary. 
One of the modern methods for human pose and appearance estimation is MonoPerfCap of Xu {et al.}~\cite{Xu2018}. 
It imposes implicit bone length constraints through a dense template tailored to a specific person and captured in an external acquisition process. 

Recently, many learning-based approaches for human pose and hand pose estimation have been presented in the literature ~\cite{RogezWS18,DBLP:conf/cvpr/KanazawaBJM18,pavlakos2018ordinal,Moreno-Noguer2017,Martinez2017,malik2019whsp,malik2018structure,malik2018deephps,malik2017simultaneous}. 
In~\cite{DBLP:conf/iccv/ZhouH0XW17}, weak supervision constrains the output of the network with fixed bone proportions taken from the training dataset. Sun {et al.}~\cite{sun17} exploit a~joint connection structure and uses bones instead of joints for pose representation. Wandt and Rosenhahn~\cite{DBLP:journals/corr/abs-1902-09868} use kinematic chain representation and include bone length information to their loss function during training. In contrast to our SfAM, \cite{DBLP:journals/corr/abs-1902-09868} is not as robust to noisy 2D input (see~Section~\ref{sec:robustness2D}). 
All~these methods are highly specialized and rely on extensive collections of training data.
In contrast, our~SfAM is a general approach that can cope with different articulated structures, with no need for labeled~datasets. 

\section{The Proposed SfAM Approach}\label{sec:method} 
\begin{figure}[H]
\begin{center}
     \includegraphics[width=13.6cm]{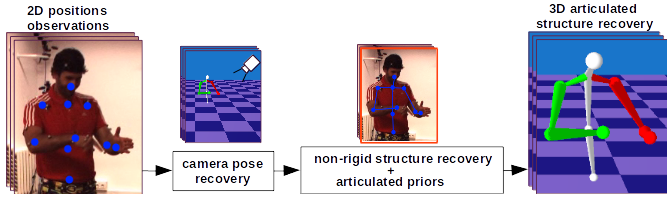}\\
   \caption{{The pipeline} of the proposed SfAM approach. Following factorization-based NRSfM, we first recover the camera pose using 2D position observations. Then, we recover 3D articulated structure by~optimizing our new energy functional accounting for articulated priors.}
\label{fig:MethodOverview} 
\end{center}
\end{figure}

Figure~\ref{fig:MethodOverview} shows a high-level overview of our approach. 
Following factorization-based NRSfM~\cite{Dai2014}, we first recover the camera pose using 2D landmarks (Section~\ref{sec:camera_poses}). 
For 3D structure recovery, we~extend the target energy function of the classic NRSfM problem~\cite{Ansari2017,Dai2014} by our articulated prior term (Section~\ref{sec:articulate_structure}). 

We assume that sparse 2D correspondences are given.
In {Section}~\ref{sec:energy_optimisation}, we show how our new energy is efficiently optimized alternating between {{fixed-point continuation algorithm}}~\cite{DBLP:journals/mp/MaGC11} and {{Levenberg--Marquardt}}~\cite{Levenberg_1944, Marquard_1963}.
 This leads to an accurate reconstruction of articulated motions of different~structures. 

\subsection{Factorization Model}\label{sec:factor_model}

The input to SfAM is the measurement matrix $ \bfW = [\bfW_1, \bfW_2, \hdots, \bfW_T]^\mathsf{T} \in \mathbb{R}^{2T\times N}$ with $N$ 2D joints tracked over $T$ frames. 
Every $\bfW_t$, $t \in \{1, \hdots, T\}$, is registered to the centroid of the observed structure and the translation is resolved in advance. 
Most of the NRSfM methods assume orthographic projection, as the intrinsic camera model is usually not known. Even though some benchmarks (e.g.,~\cite{Ionescu2014}) provide camera parameters, we develop a general approach for uncalibrated settings.
Following standard SfM approaches, we assume that every 2D projection $\bfW_t$ can be factorized into a camera pose-projection matrix $\bfR_t \in \mathbb{R}^{2 \times 3}$ and 3D structure $\bfS_t \in \mathbb{R}^{3 \times N}$ so that $\bfW_t = \bfR_t \bfS_t$. 
We assume that the~articulated structure deforms under the low-rank shape model~\cite{Bregler2000, Ansari2017}. 
Thus, $\bfS = [\bfS_1, \bfS_2, \hdots, \bfS_T]^\mathsf{T}$ can be parametrized by the set of unknown basis shapes $\bfB \in \mathbb{R}^{3K\times N}$ of cardinality $K$ and the coefficient matrix $\bfC \in \mathbb{R}^{T\times K}$: 
\begin{align}\label{eq:main} 
  \bfW = \bfR \bfS = \underbrace{\bfR \, (\bfC\otimes \bfI_3)}_{\bfM} \bfB = \bfM \bfB, 
\end{align}
where 
$\bfR = \operatorname{bkdiag}(\bfR_1, \bfR_2, \hdots, \bfR_T)$ is the joint camera pose-projection matrix, 
$\bfI_3$ is a $3 \times 3$ identity matrix and $\otimes$ denotes Kronecker product. 

\subsection{Recovery of Camera Poses}\label{sec:camera_poses} 
 
Applying singular value decomposition to $\bfW$, we obtain initial estimates of $\bfM$ and $\bfB$ from Equation \eqref{eq:main} up~to~an~invertible 
corrective transformation 
$\bfQ \in \mathbb{R}^{3K\times 3K}$: %
\begin{align}\label{eq:svdW} 
  \bfW \cong \bfM' \bfB' \cong \underbrace{\bfM' \bfQ}_{\bfM} \, \underbrace{\bfQ^{-1} \bfB'}_{\bfB}=\bfM\bfB. 
\end{align} 

In the following, we are using the shortcuts 
$\bfM'_{2t-1:2t} \in \mathbb{R}^{2\times 3K}$ for every $t$-th %
pair of rows of $\bfM$, $\bfQ_k \in \mathbb{R}^{3K\times 3}$ for the $k$-th column triplet of $\bfQ$, $k \in \{1,\hdots,K\}$. 
Considering~\eqref{eq:main} and \eqref{eq:svdW}, for every $t \in \{ 1,\hdots, T\}$ and $k \in \{ 1,\hdots, K\}$, we have: 
\begin{align}\label{eq:3} 
\bfM'_{2t-1:2t}\bfQ_k=c_{tk}\bfR_t.
\end{align}

Using the orthonormality constraints $\bfR_t \bfR_t^\mathsf{T}=\bfI_2$ and denoting $\bfF = \bfQ \bfQ^\mathsf{T}$, we obtain: 
\begin{equation}
\begin{cases}\label{eq:4} 
\bfM'_{2t-1}\bfF_k\bfM'^T_{2t-1} = \bfM'_{2t}\bfF_k\bfM'^T_{2t}=c_{ik}^2\bfI_2, \\
\bfM'_{2t-1}\bfF_k \bfM'^T_{2t} = 0. 
\end{cases}
\end{equation}

Therefore, the following systems of equations can be written for every $t$ and $k$: 
\begin{align} 
  \underbrace{ 
\begin{bmatrix} 
\bfM'_{2t-1}\otimes \bfM'^T_{2t-1} - \bfM'_{2t}\otimes \bfM'^T_{2t} \\ 
\bfM'_{2t-1}\otimes \bfM'^T_{2t} 
\end{bmatrix} 
}_{\bfG_t} 
\operatorname{vec}(\bfF_k) = 0, 
\end{align} 
where $\operatorname{vec}(\cdot)$ is vectorization operator permuting a $m\times n$ matrix to a $mn$ column vector. %
Stacking all $\bfG_t$ vertically, we obtain: 
\begin{align}\label{eq:AvecFk} 
  \bfG \operatorname{vec}(\bfF_k) = 0, 
\end{align} 
where $\bfG = [\bfG_1, \bfG_2, \hdots, \bfG_T]^\mathsf{T}$. 
\textls[-40]{Finding an optimal $\bfF_k$ can be performed by solving the~optimization~problem:}
\begin{align}\label{eq:opt_problem} 
  \underset{\;\;\bfF_k}{\min} \norm{\bfG \operatorname{vec}(\bfF_k) }^2. 
\end{align}

Due to the rank-3 constraint on every $\bfF_k$, this problem is solved by the iterative shrinkage-thresholding (IST) method~\cite{beck2009fast}. %
Once an optimal $\bfF$ is found, the corrective transformation $\bfQ$ is recovered by Cholesky decomposition. 
Using $\bfQ$, $\bfR$ is recovered from Equations~\eqref{eq:main}--\eqref{eq:4}. %

\subsection{Articulated Structure Recovery} 
\unskip
\subsubsection{Articulated Structure Representation}\label{sec:articulate_structure}

Having found $\bfR$, we recover $\bfS$. 
Note that we optionally rely on an updated $\bfW$ after the smooth shape trajectory step which  
imposes additional constraints on point trajectories and reduces the overall number of unknowns; please refer to~\cite{Ansari2017} for more details. 
We rearrange the shape matrix $\bfS$ to  
\small
\begin{equation}
\bfS^{\#} = 
\begin{bmatrix}
X_{11} \hdots X_{1N} & Y_{11} \hdots Y_{1N} & Z_{11} \hdots Z_{1N} \\ 
\vdots \quad\quad\quad \vdots & \vdots \quad\quad\quad \vdots & \vdots \quad\quad\quad \vdots \\
X_{T1} \hdots X_{TN} & Y_{T1} \hdots Y_{TN} & Z_{T1} \hdots Z_{TN}
\end{bmatrix},
\end{equation}
\normalsize
where $(X_{tn},Y_{tn},Z_{tn}), n \in \{ 1,\hdots, N\}$ is a 3D coordinate of each joint in $\bfS$. 
$\bfS^{\#}$ can be represented as: 
\begin{equation} 
\bfS^{\#}=[\bfP_x \bfP_y \bfP_z](\bfI_3 \otimes \bfS), 
\end{equation} 
where $\bfP_x, \bfP_y, \bfP_z \in \mathbb{R}^{T \times 3N}$ are  binary row selectors. 
We follow~\cite{Ansari2017,Dai2014} and represent the optimal non-rigid structure by: 
\begin{equation}
\label{eq:bmm_convex_min}
\min_\bfS ||\bfS^{\#}\boldsymbol{\Pi}||_*, 
\quad \text{s. t.} \quad \bfW = \bfR \bfS, 
\end{equation} 
where $\boldsymbol{\Pi} = (\bfI - \frac{1}{T} \boldsymbol{1}\boldsymbol{1}^\mathsf{T})$ ($\boldsymbol{1}$ is a vector of ones) and $||.||_*$ denotes the nuclear norm. 
Note that \mbox{$\operatorname{rank}(\bfS^{\#})\leq K$}, and the mean 3D component is removed from $\bfS^{\#}$. 
As shown in Figure~\ref{fig:nrsfm}, non-rigid structures recovered by the optimization of \eqref{eq:bmm_convex_min} can have significant variations in bone lengths.
This often leads to unrealistic poses and body proportions. %
Unlike general non-rigid structures, in~articulated structures, individual rigid parts or bones have constant lengths throughout the whole sequence. 
Moreover, all the bones follow constant proportions. These constraints are called \textit{{articulated priors}}. 
We incorporate the articulated priors into the objective function \eqref{eq:bmm_convex_min} in the form of the following energy term:  
\begin{equation}
\label{eq:bl_term}
\bfE_{BL}(\bfS)=\sum_{t=1}^{T}\sum_{b=1}^{B}e_{tb}(\bfS), 
\end{equation}
where $e_{tb}(\bfS)=(D_{b}^t-L_{b})^2$ is an energy term for bone $b$ and frame $t$, $L_{b}$ is initial normalized bone length value of bone $b$. The normalization is done with respect to the sum of all initial bone lengths.
\mbox{$D_{b}^t=||X_{a_b}^t-X_{c_b}^t||_2$} is Euclidian distance between joints $X_{a_b}^t$ and $X_{c_b}^t$ connected by bone $b$; $B$ is the~number of bones of the articulated structure. Vectors $a = [X_{a_1}, X_{a_2}, \hdots, X_{a_B} ]$ and $c = [X_{c_1}, X_{c_2}, \hdots, X_{c_B} ]$ define the~parent and child joints of bones, respectively.

Unlike some previous works~\cite{DBLP:conf/eccv/DabralMKASJ18, Akhter2015, Yasin2015, DBLP:conf/iccv/ZhouH0XW17}, we do not require predefined bone lengths or~proportions. SfAM recovers optimal articulated structure that minimizes the total energy: 
\begin{equation}
\label{eq:bmm_convex_min_bl}
\min_\bfS \Big( ||\bfS^{\#}||_*+\frac{\beta}{2}\bfE_{BL}(\bfS) \Big), 
\quad \text{s. t.} \quad \bfW = \bfR \bfS,
\end{equation} 
where 
$\beta$ is a scalar weight. 
Implementation of articulated prior \eqref{eq:bl_term} as a soft constraint makes the~overall method robust to incorrect initialization of bone lengths.

\subsubsection{Energy Optimization}\label{sec:energy_optimisation} 

Since \eqref{eq:bmm_convex_min_bl} contains a nonlinear term $\bfE_{BL}(\bfS)$, we introduce an auxiliary variable $\bfA$ and obtain 
the~following optimization problem which is linear with respect to $\bfS$: 
\begin{equation}
\begin{aligned}
\label{eq:bmm_convex_min_bl_sub}
&\min_\bfS ||\bfS^{\#}||_*+\frac{\beta}{2}\min_\bfA\bfE_{BL}(\bfA), \\
&\;\;\text{s. t.}\;\; \bfW = \bfR \bfS \;\text{and}\;\bfA=\bfS.
\end{aligned} 
\end{equation} 

We rewrite \eqref{eq:bmm_convex_min_bl_sub} in the Lagrangian form: 
\begin{equation} 
\begin{aligned} 
\label{eq:lagr_min_bl} 
&\bfL(\bfS,\bfA,\mu)=\mu||\bfS^{\#}||_*+ \frac{\beta}{2}\bfE_{BL}(\bfA) 
+\frac{1}{2}||\bfW-\bfR\bfS||^2_F +\frac{1}{2}||\bfA-\bfS||^2_F, 
\end{aligned} 
\end{equation} 
where $||.||_F$ denotes the Frobenius norm and $\mu$ is a parameter.
We split $\eqref{eq:lagr_min_bl}$ into two subproblems: 
\begin{equation} 
\begin{aligned} 
\label{eq:two_subproblems1} 
&\min_{\bfS}\bfL(\bfS,\mu)=&\min_{\bfS}\Big(\mu||\bfS^{\#}||_* + 
\frac{1}{2}||\bfW-\bfR\bfS||^2_F + \frac{1}{2}||\bfA-\bfS||^2_F\Big) 
\end{aligned} 
\end{equation} 
\begin{align} 
\label{eq:two_subproblems2} 
& \;\text{and}\; \min_{\bfA}\bfL(\bfA)=\min_{\bfA}\Big(\frac{\beta}{2}\bfE_{BL}(\bfA) +\frac{1}{2}||\bfA-\bfS||^2_F\Big). 
\end{align} 

We alternate between the subproblems \eqref{eq:two_subproblems1} and \eqref{eq:two_subproblems2} and iterate until convergence. 
$\bfA$ remains fixed in \eqref{eq:two_subproblems1} and $\bfS$ remains fixed in \eqref{eq:two_subproblems2}. In every optimization step, the subproblem~\eqref{eq:two_subproblems1} updates the~3D structure so that it more accurately projects to the observed 2D landmarks. The subproblem~\eqref{eq:two_subproblems2} penalizes the difference in bone lengths among all frames while recovering the sequence-specific bone proportions. The bone lengths of the recovered optimal 3D structures are almost constant throughout the whole image sequence but different from the initial $L_{b}$.

The subproblem \eqref{eq:two_subproblems1} is linear and solved by the fixed-point continuation (FPC) method~\cite{DBLP:journals/mp/MaGC11}. 
First,~we obtain the gradient of $\frac{1}{2}(||\bfW-\bfR\bfS||^2_F+||\bfA-\bfS||^2_F)$ with respect to $\bfS^{\#}$: 
\begin{equation}
\begin{aligned}
\label{eq:gradient_g}
g(\bfS^{\#},\bfA)=\frac{\partial \frac{1}{2}(||\bfW-\bfR\bfS||^2_F+||\bfA-\bfS||^2_F)}{\partial \bfS^{\#}}=
[\bfP_x \bfP_y \bfP_z](\bfI_3\otimes(\bfR^\mathsf{T}(\bfR\bfS-\bfW)+(\bfS-\bfA))). 
\end{aligned} 
\end{equation} 

Next, FPC for $\min_{\bfS}\bfL(\bfS,\mu)$ instantiates as: 
\begin{equation}
\begin{aligned}
\label{eq:two_line}
& \bfY^{(t+1)} = \bfS^{\# (t)}-\tau g(\bfS^{\# (t)},\bfA^{(t)}), \\ 
&\,\bfS^{\# (t+1)} = \mathcal{S}_{\tau\mu^{(t)}}(\bfY^{(t+1)}), \\
&\,\mu^{(t+1)} = \rho \mu^{(t)}, 
\end{aligned} 
\end{equation} 
where $\mathcal{S}_{\nu}(\cdot)$ is the matrix shrinkage operator~\cite{DBLP:journals/mp/MaGC11} and $\tau > 0$ is a free parameter.

The second subproblem \eqref{eq:two_subproblems2} is nonlinear and is optimized for each iteration \eqref{eq:two_line} using Levenberg--Marquardt of \textit{{ceres}} ~\cite{ceressolver}. 
Let denote the $r_l$, $l \in \{1, \hdots, TN\}$ residuals of $\frac{1}{2}||\bfA-\bfS||^2_F$. 
We aggregate all residuals $e_{tb}(\bfA)$ from \eqref{eq:bl_term} {(}note that $\bfS$ in \eqref{eq:bl_term} is substituted by $\bfA${)} 
 and $r_l$ into a~single~function: 
\begin{equation} 
\begin{aligned}\label{eq:F_vector_valued} 
\bfF(\bfA) = &[e_{11}(\bfA), \hdots, e_{BT}(\bfA), r_1, \hdots, r_{TN}]^\mathsf{T} : \\ &\mathbb{R}^{3TN} \to \mathbb{R}^{BT + TN}. 
\end{aligned} 
\end{equation} 

Next, the objective function \eqref{eq:two_subproblems2} can be compactly written in terms of $\bfA$ as:  
\begin{align} 
\bfL(\bfA) = \norm {\bfF(\bfA)}_2^2. 
\end{align} 

The target nonlinear energy optimization problem 
consists of finding an optimal parameter set $\bfA'$ so that: 
\begin{align}\label{eq:nl_problem} 
\bfA' = \arg\min_{\bfA} \norm{\bfF(\bfA)}_2^2. 
\end{align} 

We solve \eqref{eq:nl_problem} iteratively. 
In every optimization step \textit{k}, the objective is linearized in the vicinity of~the~current solution $\bfA_k$ by the first-order Taylor expansion: 
\begin{align} 
\bfF(\bfA_k + \Delta \bfA) \approx \bfF(\bfA_k) + \bfJ(\bfA_k) \Delta \bfA, 
\end{align}
with $\bfJ(\bfA)_{(BT + TN) \times 3TN}$ being the Jacobian of $\bfF(\bfA_k)$. 
For every iteration, the objective for $\Delta \bfA$ reads:  
\begin{align} 
\min_{\Delta \bfA} \norm{\bfJ(\bfA_k) \Delta \bfA + \bfF(\bfA_k)}^2. 
\end{align} 
 
\textls[-15]{In \textit{{ceres}}~\cite{ceressolver}, the optimum is computed in the least-squares sense with the Levenberg--Marquardt~method: }
\begin{align} 
[\bfJ(\bfA_k)^\mathsf{T} \bfJ(\bfA_k)+\lambda_k \bfI]\,\Delta \bfA = -\bfJ(\bfA_k)^\mathsf{T} \bfF(\bfA_k), 
\end{align} 
where $\lambda_k > 0$ is a parameter and $\bfI$ is an identity matrix. 

The algorithm is summarized in Algorithm~\ref{algorithm:energy_optim}. 
\begin{algorithm}[H]
  \caption{Structure from Articulated Motion (SfAM)} 
  \begin{algorithmic}
  \STATE \textbf{Input:} initial normalized bone lengths $L_b$, measurement matrix $\bfW \in \mathbb{R}^{2T \times N}$ with 2D point tracks\\
  \textbf{Output:} poses $\bfR \in \mathbb{R}^{2T \times 3T}$ and 3D shapes $\bfS \in \mathbb{R}^{3T \times N}$ \\ 
  \textbf{Initialize:} $\bfS^{(0)}$ is initialized as in~\cite{Ansari2017}, $\bfA^{(0)}=\bfS^{(0)}$, $\beta=1.5$, $\mu^{(0)}=1$, $\rho=0.25$, $\tau=0.2$ \\
  \textbf{step 1:} recover $\bfR$ with IST method~\cite{beck2009fast} (Section~\ref{sec:camera_poses}) \\ 
  \textbf{step 2 (optional):}  smooth point trajectories in $\bfW$~\cite{Ansari2017} \\ 
  \textbf{step 3: while not converged do} \\
  $\;\;\;\;$1: $\bfA^{(t+1)}=\arg\min_{\bfA}(\frac{\beta}{2}\bfE_{BL}(\bfA) +\frac{1}{2}||\bfS^{(t)}-\bfA||^2_F$) \\
  $\;\;\;\;$(optimize with Levenberg--Marquardt~\cite{Levenberg_1944, Marquard_1963})\\
  $\;\;\;\;$2: $g^{(t+1)}=\bfR^\mathsf{T}(\bfR\bfS^{(t)}-\bfW)+(\bfS^{(t)}-\bfA^{(t+1)})$\\
  $\;\;\;\;$3: $\bfY^{(t+1)}=\bfS^{(t)}-\tau g^{(t+1)}$\\
  $\;\;\;\;$4: $\bfS^{(t+1)}=\mathcal{S}_{\tau\mu^(t)}(\bfY^{(t+1)})$\\
  $\;\;\;\;$5: $\mu^{(t+1)}=\mu^{(t)}\rho$\\
  \textbf{end while} 
  \label{algorithm:energy_optim} 
\end{algorithmic} 
\end{algorithm} 
%
\section{Experiments and Results}\label{sec:experiment} 
We extensively evaluate our SfAM on several datasets including %
Human 3.6m~\cite{Ionescu2014}, synthetic sequences of Akhter {et al.}~\cite{Akhter2011} and NYU hand pose~\cite{tompson14tog} dataset. 
Moreover, we demonstrate qualitative results on challenging community videos.
In total, our SfAM is compared to over thirty state-of-the-art model-based and learning-based methods (see Tables \ref{tab:protocol1} and \ref{tab:Synthetic}).  
We also implement {SMSR} 
of Ansari~{et~al.}~\cite{Ansari2017}, which is the most related approach to our SfAM and evaluate it on~\cite{Ionescu2014, tompson14tog} as well as community videos. 
Moreover, we extend SMSR~\cite{Ansari2017} with the local rigidity constraint of Rehan~{et al.}~\cite{Rehan2014} and include it into our comparison.

In Section~\ref{sec:robustness2D}, we evaluate the robustness of our approach to inaccuracies in 2D landmarks. 
The~proposed SfAM recovers correct articulated structures given highly inaccurate initial bone lengths in Section~\ref{sec:robustnessbonelengths}.
Finally, in Section~\ref{sec:evalhumanyoutube}, we highlight the numerous cases when our method performs better than state-of-the-art learning-based approaches in real-world scenes.

In all experiments, we use a sliding time window of $200$ frames. 
For sequences shorter than $200$~frames, we run our method on the whole sequence at once. 
All experiments are performed on~a~system with 32 GB RAM and twelve-core Intel Xeon CPU running at 3.6 GHz. 
Our framework is implemented in C++. 
Average processing time for a single frame from the Human 3.6m dataset~\cite{Ionescu2014} with given 2D annotations amounts to $140$ ms. 

\subsection{Evaluation Methodology} 

We follow the established evaluation methodology in the area of NRSfM and rigidly align our 3D reconstructions to the ground truth. 
We report the reconstruction error $\mathcal{E}_{3D}$ in mm between ground truth joint positions $\overline{\bfS_n^t}$ and aligned 3D reconstructions $G(\bfS_n^t)$: 
\begin{align}\label{req_err}
\mathcal{E}_{3D}=\min_G\frac{1}{T}\frac{1}{N}\sum_{t=1}^{T}\sum_{n=1}^{N}||\overline{\bfS_n^t}-G(\bfS_n^t)||_2,
\end{align}
where $n \in \{ 1,\hdots, N\}$, $t \in \{ 1,\hdots, T\}$, $T$ is the number of frames in the sequence and $N$ is the number of joints of the articulated object. 
For some datasets, we report the normalized mean 3D error:
\begin{equation}\label{norm_err}
\begin{aligned}
  &e_{3D}=\min_G\frac{1}{\sigma T}\frac{1}{N}\sum_{t=1}^{T}\sum_{n=1}^{N}||\overline{\bfS_n^t}-G(\bfS_n^t)||^2_2, \,\text{with} \\ 
  &\;\;\;\;\;\;\;\;\;\;\;\sigma=\min_G\frac{1}{3 T}\sum_{t=1}^{T}(\sigma_{tx}+\sigma_{ty}+\sigma_{tz}),
\end{aligned}
\end{equation}
where $\sigma_{tx}, \sigma_{ty}$ and $\sigma_{tz}$ denote normalized variances of reconstructions $G(\bfS_n^t)$ along the~$x,y,z$-axes~respectively. 

\subsection{Human Pose Estimation\label{sec:evalhumanpose} }
\unskip
\subsubsection{Human 3.6m Dataset \label{sec:evalhuman} }
\textbf{{Human 3.6m}}~\cite{Ionescu2014} is currently the largest dataset for monocular 3D human pose sensing. It~is widely used for evaluation of learning-based human pose estimation methods. 
Table~\ref{tab:protocol1} gives an~overview of the quantitative results on the Human 3.6m~\cite{Ionescu2014}. 
We highlight approaches that are trained on Human 3.6m~\cite{Ionescu2014} with ``*''. 
We follow three common evaluation protocols. 
In \textbf{{Protocol} \#1}, we compare the methods on two subjects ($S9$ and $S11$). The original framerate $50$~$fps$ is reduced to $10$~$fps$. 
The learning-based approaches marked with ``*'' use subjects $S1$, $S5$, $S6$, $S7$, $S8$ and all camera views for training.
Testing is done for all cameras. 
For \textbf{{Protocol} \#2}, only the frontal view (``camera3'') is used for evaluation. 
For \textbf{{Protocol} \#3}, evaluation is done on every $64^{th}$ frame of subject $S11$ for all cameras. 
The learning-based approaches marked with ``*'' use subjects $S1$, $S5$, $S6$, $S7$, $S8$ and $S9$ for~training. 

For all methods and under all evaluation protocols, we report the reconstruction error $\mathcal{E}_{3D}$ after the rigid alignment of the recovered structures with ground truth. 
In our method, the bone lengths are initialized with the average values for all the subjects from the dataset.

As we see from Table~\ref{tab:protocol1}, we show competitive accuracy to best performing learning-based approaches that are trained on Human 3.6m~\cite{Ionescu2014}. In Section~\ref{sec:evalhumanyoutube}, we demonstrate that our approach works better in real-world scenes which are different from this dataset. %
\begin{table}[H]
\centering
\caption{{The reconstruction} error $\mathcal{E}_{3D}$ of {SfAM} and previous 
methods on Human 3.6m dataset. ``*'' indicates learning-based methods which are trained on Human 3.6m~\cite{Ionescu2014}. 
We outperform all~model-based approaches and reach very close to the tuned supervised learning techniques.}
\setlength{\tabcolsep}{15pt}
\begin{tabular}{@{}l c c c c@{}}
\toprule
\textbf{Method}&\textbf{P1}&\textbf{P2}&\textbf{P3}\\
\midrule
{Zhou {et al.}~\cite{zhou2016sparseness} *} & 106.7&- & - \\
{Akhter {et al.}~\cite{Akhter2015}} &-& 181.1 &-  \\
{Ramakrishna {et al.}~\cite{Ramakrishna2012}} &-& 157.3 &-  \\
{Bogo {et al.}~\cite{Bogo2016}} &-& 82.3&-   \\
{Kanazawa {et al.}~\cite{DBLP:conf/cvpr/KanazawaBJM18} *} & 67.5 & 66.5&-\\
{Moreno-Noguer~\cite{Moreno-Noguer2017} *} & 62.2 & - & -\\
{Yasin {et al.}~\cite{Yasin2015}} &-&-& 110.2 \\
{Rogez {et al.}~\cite{Rogez2016}} &-&-& 88.1 \\
{Chen, Ramanan~\cite{Chen2016} *} &-&-& 82.7 \\
{Nie {et al.}~\cite{Nie2017} *} &-&- & 79.5 \\
{Sun {et al.}~\cite{sun17} *} &-&-& 48.3\\
{Omran {et al.}~\cite{DBLP:conf/3dim/OmranLPGS18} *} & 59.9 & - & -\\
{Zhou {et al.}~\cite{DBLP:journals/pami/ZhouZPLDD19} *} & 54.7 & - & -\\
{Mehta {et al.}~\cite{DBLP:conf/3dim/MehtaRCFSXT17} *} & 54.6 & - & -\\
{Pavlakos {et al.}~\cite{pavlakos2017} *} & 51.9& - & - \\
{Kinauer {et al.}~\cite{Kinauer17} *} & 50.3& - & -\\
{Tekin {et al.}~\cite{TekinMSF17} *} & 50.1& - & - \\
{Rogez {et al.}~\cite{RogezWS18} *} & 49.2& 51.1&42.7\\
{Habibie {et al.}~\cite{DBLP:journals/corr/abs-1904-03289} *} & 49.2& - & -\\
{Martinez {et al.}~\cite{Martinez2017} *} & 45.6& - & -\\
{Zhao {et al.}~\cite{DBLP:journals/corr/abs-1904-03345} *} & 43.8& - & -\\
{Pavlakos {et al.}~\cite{pavlakos2018ordinal} *} & 41.8& - & -\\
{Arnab, Doersch {et al.}~\cite{DBLP:journals/corr/abs-1905-04266} *} & 41.6& - & -\\
{Chen, Lin {et al.}~\cite{DBLP:journals/corr/abs-1903-08839} *} & 41.6& - & -\\
{Sun {et al.}~\cite{DBLP:conf/eccv/SunXWLW18} *} & 40.6& - & -\\
{Wandt, Rosenhahn~\cite{DBLP:journals/corr/abs-1902-09868} *} & 38.2& - & -\\
{Pavllo {et al.}~\cite{DBLP:journals/corr/abs-1811-11742} *} & 36.5& - & -\\
{Dabral {et al.}~\cite{DBLP:conf/eccv/DabralMKASJ18} *} & 36.3& - & -\\
{SMSR~\cite{Ansari2017}} & 106.6 & 105.2 & 102.9\\
{{SMSR}~\cite{Ansari2017}+\cite{Rehan2014}} & 145.2 & 124.0 & 139.9\\ 
{Our SfAM} & 51.2 & 51.7 & 53.9\\
\bottomrule
\end{tabular}
\label{tab:protocol1}
\end{table}
\unskip
\begin{table}[H]
\centering
\caption{{The normalized} mean 3D error $e_{3D}$ of previous {NRSfM} methods 
and our SfAM for synthetic sequences~\cite{Akhter2008}.}  
\begin{tabular}{c c c c c}
\toprule
\textbf{Method} & \textbf{Drink} & \textbf{PickUp} & \textbf{Stretch} &\textbf{Yoga}\\
\midrule
{MP~\cite{Paladini2009}} & 0.4604 & 0.4332 & 0.8549 & 0.8039 \\
{PTA~\cite{Akhter2008}} & 0.0250 & 0.2369 & 0.1088 & 0.1625 \\
{CSF1~\cite{Gotardo2011}} & 0.0223 & 0.2301 & 0.0710 & 0.1467 \\
{CSF2~\cite{GotardoM2011}} & 0.0223 & 0.2277 & \textbf{{0.0684}} & 0.1465\\
{BMM~\cite{Dai2014}} & 0.0266 & \textbf{{0.1731}} & 0.1034 & \textbf{{0.1150}}\\
{Lee~\cite{Lee2016}} & 0.8754 & 1.0689 & 0.9005 & 1.2276\\
{PPTA~\cite{Agudo_etal_cviu2018}} & \textbf{0.011} & 0.235 & 0.084 & 0.158\\
{SMSR~\cite{Ansari2017}} & 0.0287 & \textbf{0.2020} & 0.0783 & 0.1493\\
{{SMSR}~{\cite{Ansari2017}+\cite{Rehan2014}}}& 0.4348 & 0.4965  & 0.3721  & 0.4471  \\
{Our SfAM}  &  0.0226 & \textbf{0.1921}  & \textbf{0.0673}  & \textbf{0.1242 }  \\
\bottomrule
\end{tabular} 
\label{tab:Synthetic} 
\end{table}

\begin{figure}[H]
\centering
\includegraphics[width=12.6cm]{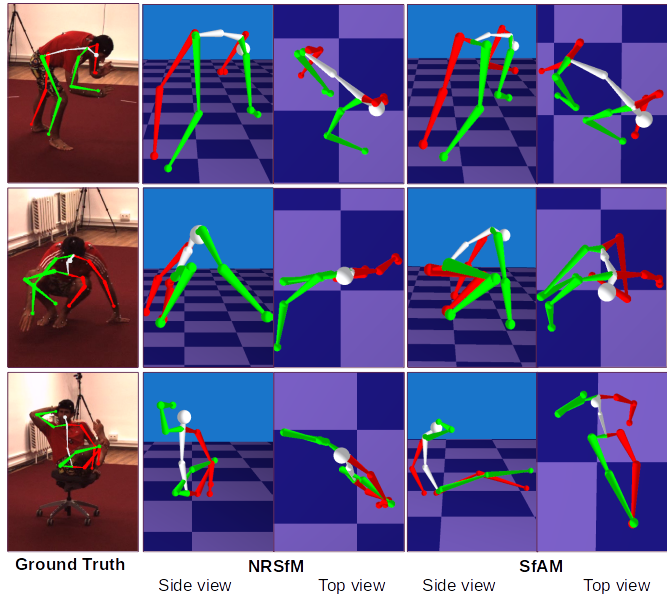}\\
   \caption{Comparison of our SfAM and NRSfM~\cite{Ansari2017} on Human 3.6m~\cite{Ionescu2014}. NRSfM considers humans as general non-rigid objects 
   and changes bone lengths from frame to frame. 
   }
   \label{fig:human_seq}
\end{figure}

In Figure~\ref{fig:human_seq}, we~visualize several reconstructions of highly challenging scenes by SMSR~\cite{Ansari2017} and the proposed SfAM. See Figure~\ref{fig:appendix} for additional visualizations.

\subsubsection{Robustness to Inaccurate 2D Point Tracks}\label{sec:robustness2D} 

We validate the robustness of our approach to inaccuracies in 2D landmarks on Human 3.6m~\cite{Ionescu2014}. 
We compare our SfAM to state-of-the-art learning-based methods~\cite{DBLP:journals/corr/abs-1902-09868,Moreno-Noguer2017,Martinez2017} trained on ground truth 2D data. 
We add Gaussian noise with increasing values of the standard deviation to the  2D ground truth point tracks. 
The reconstruction error as the function of the standard deviation of the noise is plotted in Figure~\ref{fig:Plot_BL_Noise}a. 
SfAM is more robust than the compared methods for moderate and high perturbations, and the error grows very slowly with the increasing noise level. 
In contrast to our SfAM, the errors of~\cite{DBLP:journals/corr/abs-1902-09868,Moreno-Noguer2017,Martinez2017} grow very fast even with a low level of noise. 
Note that we evaluate our method on~a~higher level of noise than~\cite{DBLP:journals/corr/abs-1902-09868,Moreno-Noguer2017,Martinez2017}. 
The average error of the currently best performing 2D detectors is between 10--15 pixels~\cite{Wei2016ConvolutionalPM, stackedhourglass}. We see that, for 10--15 pixels, SfAM has comparable error to the most accurate learning-based approaches while not relying on training data and being generalizable for different object classes.

\subsubsection{Robustness to Incorrectly Initialized Bone Lengths and Real Bone Length Recovery} \label{sec:robustnessbonelengths}

\begin{figure}[H]
\begin{center}
\captionsetup[subfigure]{labelformat=empty}
    \begin{subfigure}[b]{.55\linewidth}
    \includegraphics[width=7cm]{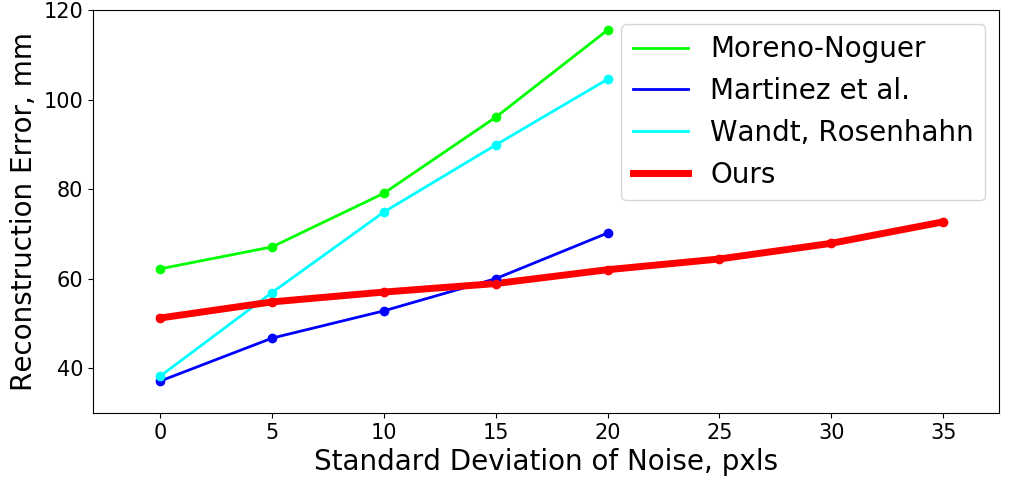}
    \centering
    \vspace{-2mm}
    \caption{(\textbf{a})\vspace{1mm}}
    \vspace{1mm}
    \end{subfigure}
    \begin{subfigure}[b]{.4\linewidth}
     \includegraphics[width=5cm]{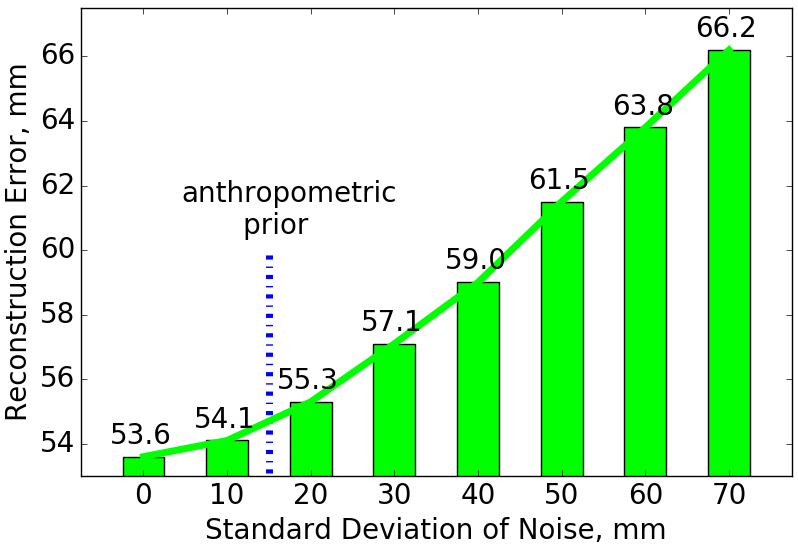}
     \centering
     \vspace{-2mm}
     \caption{(\textbf{b})\vspace{1mm}}
     \vspace{1mm}
     \end{subfigure}
    \begin{subfigure}[b]{.55\linewidth}
     \includegraphics[width=7cm]{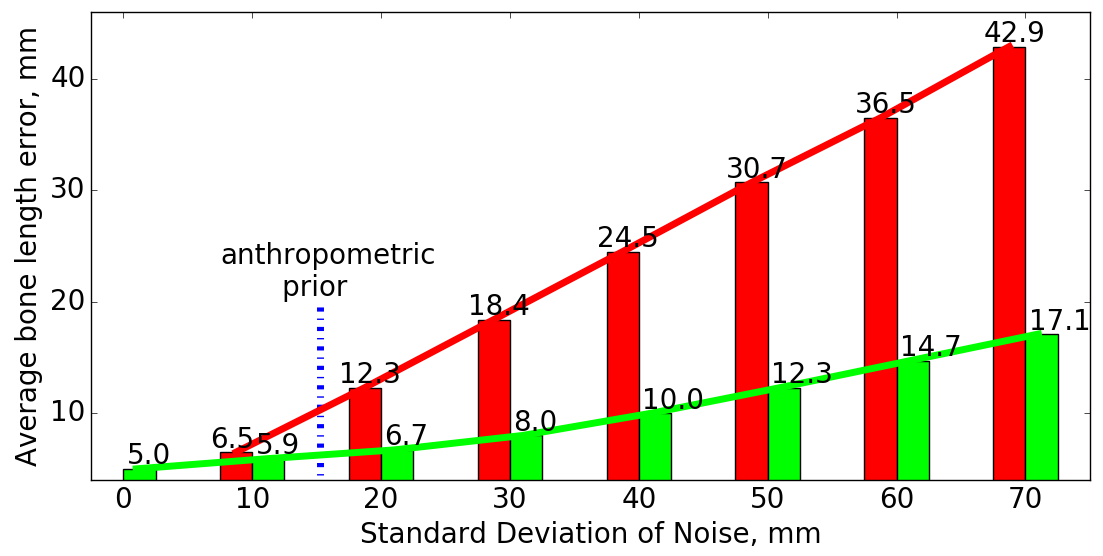}
     \centering
   \vspace{-2mm}
     \caption{(\textbf{c})}
     \end{subfigure}
     \begin{subfigure}[b]{.4\linewidth}
     \includegraphics[width=5.5cm]{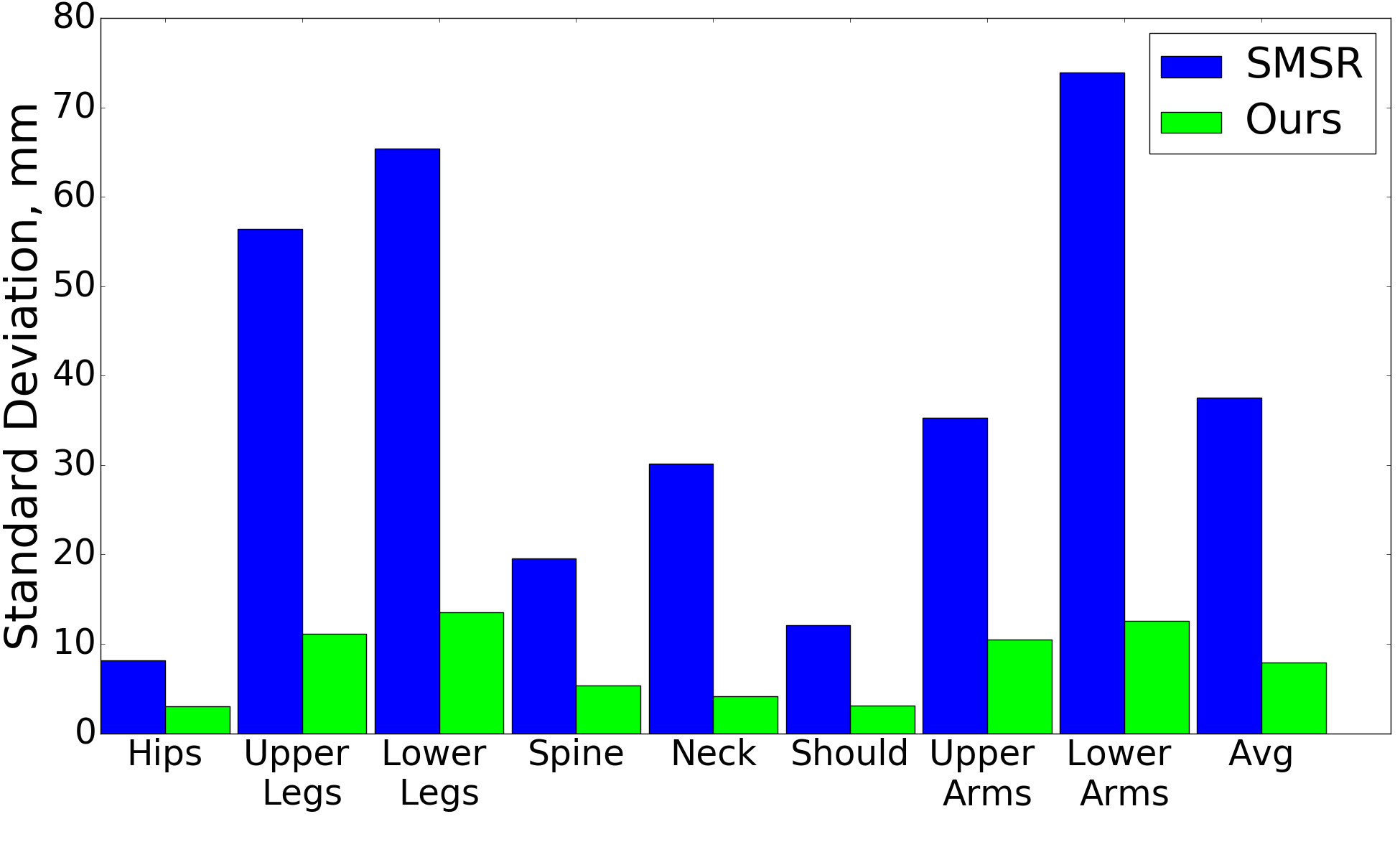}
     \centering
   \vspace{-2mm}
     \caption{(\textbf{d})}
     \end{subfigure}
   \caption{(\textbf{a}): the reconstruction error $e_{3D}$ under 2D noise; %
   (\textbf{b}): $e_{3D}$ under incorrect bone lengths initializations; 
   (\textbf{c}): average bone lengths error 
   for the increasing levels of Gaussian noise before (red) and after (green) the optimization; 
   (\textbf{d}): standard deviation of bone lengths for {SMSR}~\cite{Ansari2017} and our~SfAM. %
   }
   \label{fig:Plot_BL_Noise}
\end{center}
\end{figure}

We study the accuracy of SfAM in recovering articulated structures given incorrectly initialized bone proportions (normalized bone lengths) on the subject $S11$ from Human 3.6m~\cite{Ionescu2014}. 
Starting from the ground truth initialization of bone lengths (obtained from the dataset), we change every bone length by adding different amounts of Gaussian noise with increasing standard deviations in~the~range $[0; 70]$ mm. 
This allows us to analyze the recovered bone lengths and the robustness of SfAM to noise in a controlled and well-defined setting. 
The results of the experiment are plotted in Figure~\ref{fig:Plot_BL_Noise}b. 
If~the~structure is initialized with anthropometric priors from~\cite{anthropometricdata}, the error increases by only 3\%. 
Note~that our error in bone length estimation is  slightly affected by the increasing levels of noise. 
It~is~equal to $54$ mm with ground truth initialization and grows just to $66$ mm with $\sigma = 70$ mm. 
Note that the anthropometric prior corresponds to $\sigma \approx 15$ mm. 

Given incorrect initial bone lengths, SfAM recovers not only correct poses, but also accurate sequence-specific bone lengths. We calculate the average difference between ground truth bone lengths of subject $S11$ and the initial ones, provided to our method. We do the same for the recovered structures. 
The results are best viewed in Figure~\ref{fig:Plot_BL_Noise}c. Thus, SfAM can be used for precise skeleton estimation. %

We also calculate standard deviations of bone lengths of the reconstructed objects for SMSR~\cite{Ansari2017} and SfAM. 
Figure~\ref{fig:Plot_BL_Noise}d shows that the standard deviation of bone lengths is very high for SMSR~\cite{Ansari2017}, as~it considers a human as a general non-rigid object and changes the bone lengths from frame to frame. SfAM reduces the average standard deviation by $514$\% leading to a more accurate pose reconstruction and structure recovery. In Figure~\ref{fig:Plot_BL_Noise}d, ``Upper Legs'' and ``Lower Legs'' denote bones between the~hip/knee and knee/ankle, respectively;  ``Upper Arms'' and  ``Lower Arms'' denote bones between shoulder/elbow and elbow/wrist, respectively. 

\subsubsection{Synthetic NRSfM Datasets}\label{sec:evalhumansynthetic} 
\textbf{{Synthetic sequences}} of Akhter {et al.}~\cite{Akhter2011} are commonly used for the evaluation of sparse NRSfM. 
We compare our approach with previous SfM methods on challenging synthetic sequences with a large variety of human motions \textit{{Drink}}, \textit{{Pickup}}, \textit{{Stretch}}, and \textit{{Yoga}}~\cite{Akhter2008}. 
Some pairs of joints remain locally rigid in these sequences.  
We activate the articulated constraint for those points and evaluate our method. 
Table \ref{tab:Synthetic} shows the results of SfAM and previous SfM methods. 

The errors $e_{3D}$ for other listed methods are taken from {PPTA}~\cite{Agudo_etal_cviu2018}  
and SMSR~\cite{Ansari2017}. %
Only PPTA~\cite{Agudo_etal_cviu2018} outperforms SfAM on \textit{Drink}, whereas 
CSF2~\cite{GotardoM2011} achieves a comparable $e_{3D}$.  
SfAM achieves the most consistent performance among all compared algorithms. 

\subsubsection{Real-World Videos}\label{sec:evalhumanyoutube}

\begin{figure}[H]
\begin{center}
     \includegraphics[width=13.6cm]{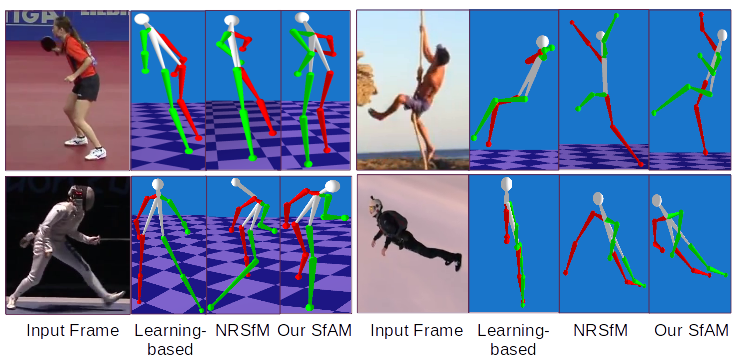} 
 \caption{Comparison of our SfAM, NRSfM~\cite{Ansari2017}, and the learning-based method of Martinez {et al.}~\cite{Martinez2017} on challenging real-world videos.} 
   \label{fig:YouTube} 
   \end{center}
\end{figure}

Our algorithm is capable of recovering human motion from challenging real-world videos. We~compare our results with the state-of-the-art learning-based approach of Martinez {et al.}~\cite{Martinez2017} and one of the best performing general-purpose NRSfM methods SMSR~\cite{Ansari2017}. 
Since ground truth 2D annotations are not available, we use OpenPose~\cite{cao2017realtime} for 2D human body landmark extraction. Bone lengths are initialized with the values from \textit{{anthropometric data tables}}~\cite{anthropometricdata}. 
As Figure~\ref{fig:YouTube} shows, ~\cite{Martinez2017} fails to correctly recover poses that are different from the training dataset~\cite{Ionescu2014}. SMSR~\cite{Ansari2017} produces unrealistic human body structures. In contrast to ~\cite{Martinez2017,Ansari2017}, our method successfully recovers 3D human poses in~real-world~scenes. 
\vspace{-2mm}
\vspace{-5mm}
\subsection{Hand Pose Estimation}\label{sec:evalhand}

\begin{figure}[H]
\begin{center}
     \includegraphics[width=13.6cm]{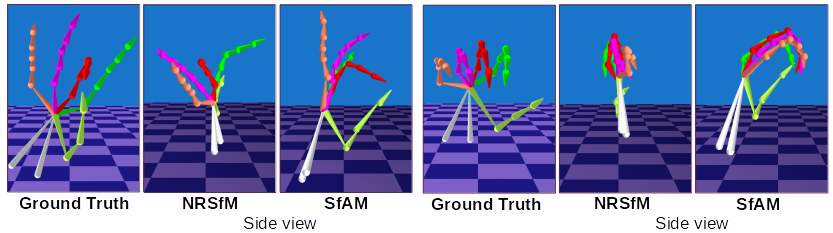}\\
   \caption{Comparison of our SfAM to NRSfM~\cite{Ansari2017} on an NYU hand pose dataset~\cite{tompson14tog}.}
   \label{fig:NYU}
\end{center}
\end{figure}

We also evaluate SfAM on the NYU hand pose dataset~\cite{tompson14tog}, which provides 2D and 3D ground truth annotations for $8252$ different hand poses. %
The hand model consists of $30$ bones. 
Hand pose recovery is a challenging problem due to occlusion and many degrees of freedom. 
We compare the~performance of our approach with SMSR~\cite{Ansari2017} and its modification with local rigidity constraint from Rehan~{et al.}~\cite{Rehan2014}. 
Quantitatively, SfAM achieves $\mathcal{E}_{3D}$ of $14.2$~{mm}. In contrast,
$\mathcal{E}_{3D}$ of SMSR~\cite{Ansari2017} is $22.2$~{mm}, and SMSR with articulated body constraints~\cite{Rehan2014} shows $\mathcal{E}_{3D}$ of $19.4$~{mm}. 
Hence, the inclusion of our articulated prior term to 
\cite{Ansari2017} achieves an error improvement of 56\%.  
The qualitative results are shown in Figure~\ref{fig:NYU}. 
Similar to human bodies, 
SfAM achieves lower error due to keeping bone lengths constant between frames. 
When SMSR~\cite{Ansari2017} fails to reconstruct the correct 3D pose, SfAM still outputs plausible results.
\vspace{-2mm} 
%

\newpage 
\section{Conclusions}\label{sec:conclusion}
\vspace{-2mm}
We present a new method for 3D articulated structure recovery from 2D landmarks. The proposed approach is general and not restricted to specific structures or motions. Integration of our soft articulated prior term into a general-purpose NRSfM approach and alternating optimization 
resulted in~accurate and stable results. 

In contrast to the vast majority of state-of-the-art approaches, SfAM does not require training data or known bone lengths. By ensuring consistency of bone lengths throughout the whole sequence, it~optimizes sequence-specific bone proportions and recovers 3D structures. In extensive experiments, it~proves its generalizability and shows accuracy close to state-of-the-art on public benchmarks. It also shows a remarkable improvement in accuracy compared to other model-based approaches. Moreover, our method outperforms learning-based approaches in complicated real-world videos. 
All in all, we~show that high accuracy on benchmarks can be achieved without the need for training and parameter tuning for specific datasets. 

In future work, we plan to apply SfAM to animal shape estimation and recovery of personalized human skeletons. 
We also believe it can 
boost the development of methods for human and hand pose estimation with semi-supervision. 


\vspace{6pt}

\funding{This research was funded by the project VIDETE of the German Federal Ministry of Education and Research (BMBF), Grant No.~01IW18002.}

\abbreviations{The following abbreviations are used in this manuscript:\\

\noindent 
\begin{tabular}{@{}ll}
SfAM & Structure from Articulated Motion\\
SfM & Structure from Motion\\
NRSfM & Non-Rigid Structure from Motion\\
FPC & Fixed-Point Continuation\\
SMSR & Scalable Monocular Surface Reconstruction\\
IST & Iterative Shrinkage-Thresholding
\end{tabular}}

\appendixtitles{no} 
\appendix
\section{}
\unskip
\vspace{-5mm}
\begin{figure}[H] 
\begin{center} 
     \includegraphics[width=14cm]{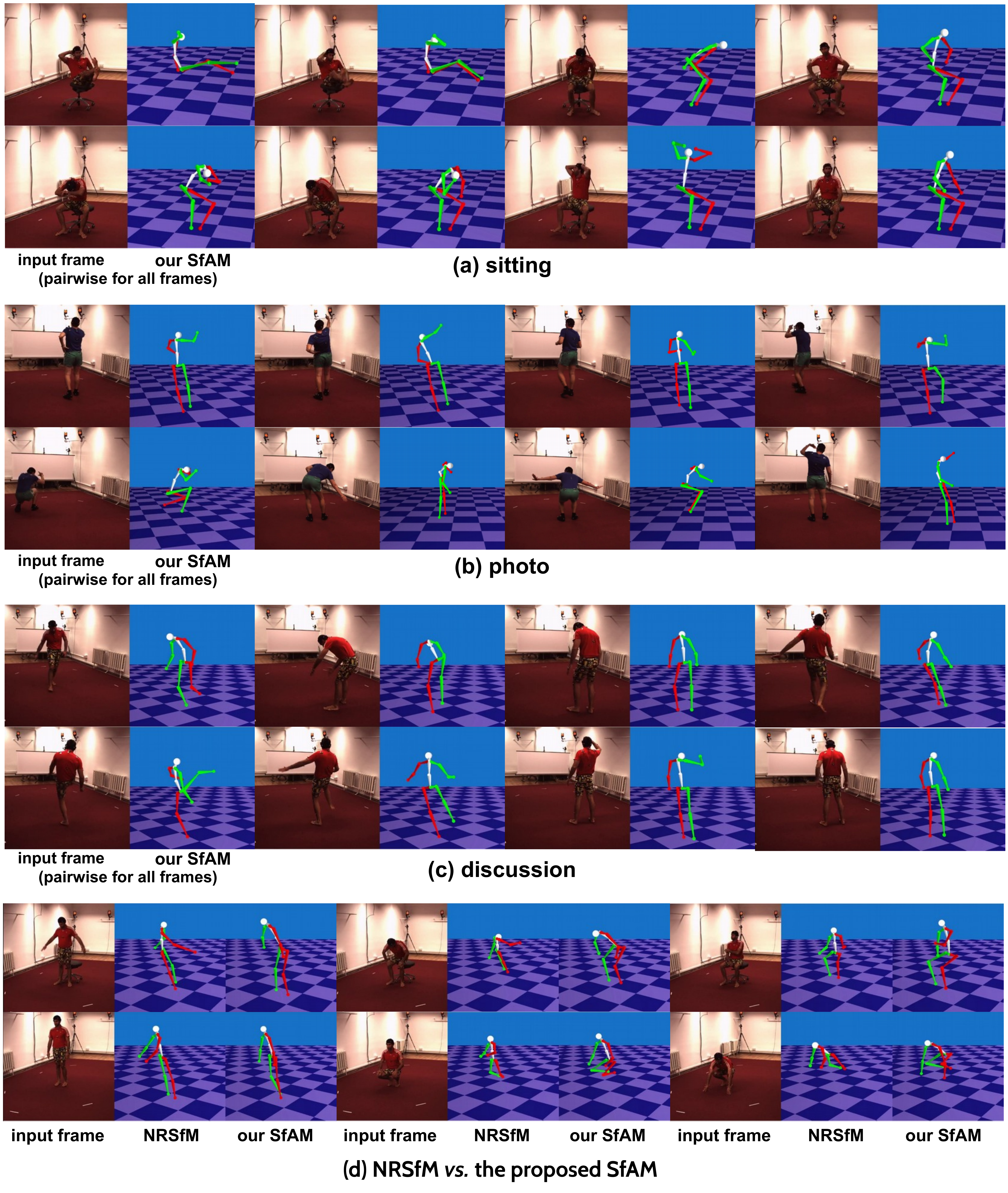} 
     \vspace{-2mm}
   \caption{Additional visualizations of our results and reconstructions with NRSfM of Ansari {et al.}~\cite{Ansari2017} on 
   several sequences from~\cite{Ionescu2014}. 
   (\textbf{a})--(\textbf{c}): our results on \textit{{sitting}}, \textit{{photo}} and \textit{{discussion}}. These sequences and poses are among the most challenging in the dataset. 
   (\textbf{d}): comparison of our SfAM and NRSfM~\cite{Ansari2017}. 
   } 
\label{fig:appendix} 
\end{center} 
\end{figure} 
\reftitle{References}

\end{document}